\documentclass{article}
\usepackage{spconf,amsmath,graphicx}
\usepackage{amsfonts}

\title{Audio-video Emotion Recognition in the Wild using Deep Hybrid Networks}
\name{Xin Guo$^{\star}$ \qquad Luisa F. Polan\'{i}a$^{\dagger}$ \qquad Kenneth E. Barner$^{\star}$\thanks{The work is supported by the National Science Foundation under Grant No. 1319598.}}

\address{$^{\star}$ University of Delaware, Department of Electrical and Computer Engineering,  Newark, DE, USA \\
\texttt{\{guoxin, barner\}@udel.edu}\\
    $^{\dagger}$Target Corporation, Sunnyvale, CA, USA\\
\texttt{Luisa.PolaniaCabrera@target.com}}

\begin{document}
%
\maketitle
\begin{abstract}
This paper presents an audiovisual-based emotion recognition hybrid network. While most of the previous work focuses either on using deep models or hand-engineered features extracted from images, we explore multiple deep models built on both images and audio signals. Specifically, in addition to convolutional neural networks (CNN) and recurrent neutral networks (RNN) trained on facial images, the hybrid network also contains one SVM classifier trained on holistic acoustic feature vectors, one long short-term memory network (LSTM) trained on short-term feature sequences extracted from segmented audio clips, and one Inception(v2)-LSTM network trained on image-like maps, which are built based on short-term acoustic feature sequences. Experimental results show that the proposed hybrid network outperforms the baseline method by a large margin. 

\end{abstract}

\section{Introduction}
\label{sec:intro}
Emotion recognition is relevant in many computing areas that take into account the affective state of the user, such as human-computer interaction~\cite{Cowi01}, human-robot interaction~\cite{Kuli07}, music and image recommendation~\cite{Shan09}, affective video summarization~\cite{Joho09}, and personal wellness and assistive technologies~\cite{Pant07}. Although emotion recognition is an interesting problem, it is also very challenging unless the recording conditions are well controlled. Emotion recognition ``in the wild" suffers from many issues that need to be overcome, such as cluttered backgrounds, large variances in face pose and illumination, video and audio noise, and occlusions.

\begin{figure*}
\centering
\includegraphics[width=0.9\textwidth]{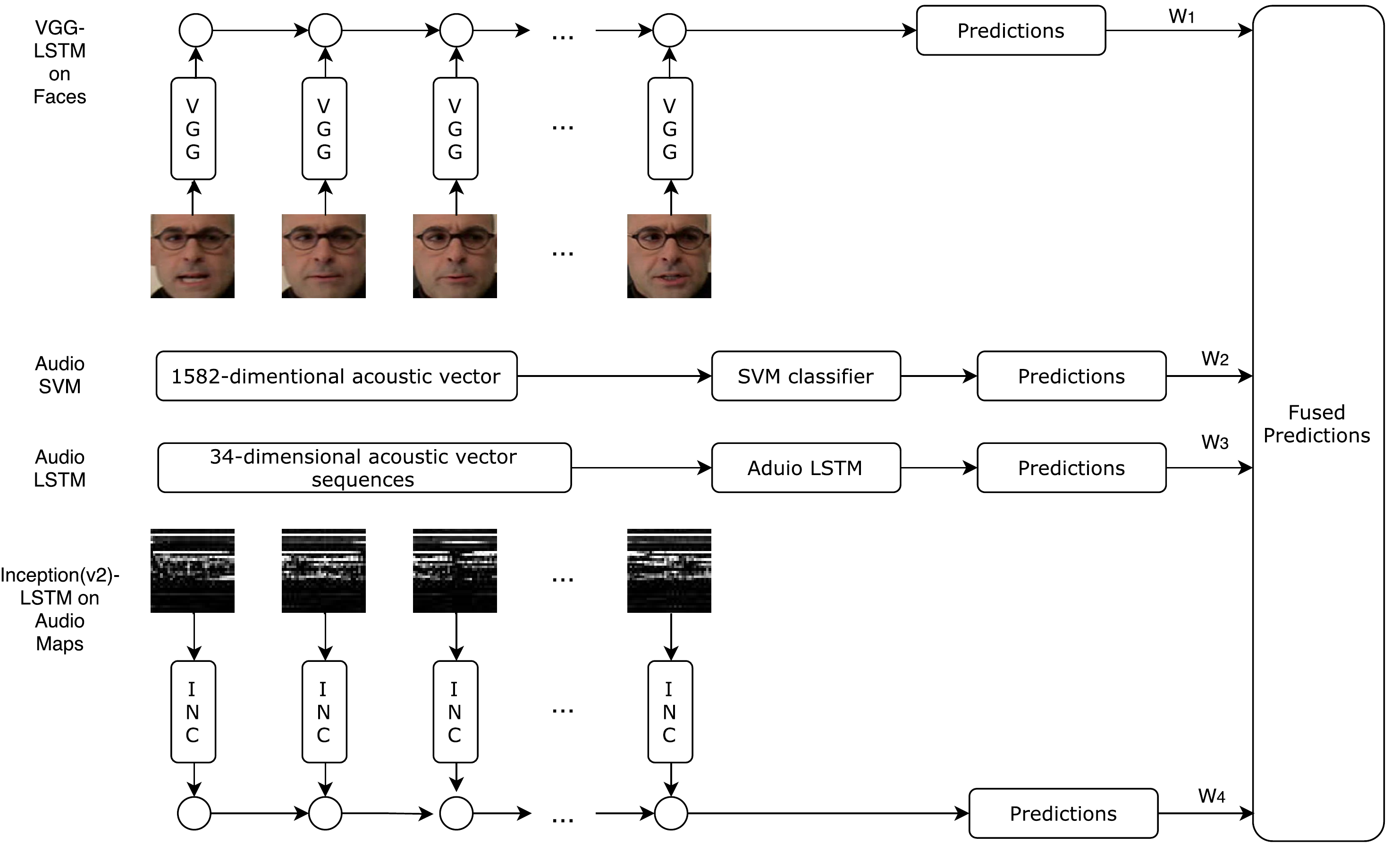}
\caption{The overall structure of the proposed hybrid network. For visualization purposes, only one VGG-LSTM on faces is shown in the diagram; however, note that the hybrid network contain two VGG-LSTMs with the same network structure but trained on faces detected by different methods.}
\label{fig1}
\end{figure*}

Recently, hybrid neural networks combining CNNs and RNNs~\cite{Fan16, Khor16, Ebra15} have become the state-of-the-art for emotion recognition. Of particular interest are the top-performing works of the EmotiW Challenge, whose goal is to advance emotion recognition in unconstrained conditions by providing researchers with a platform to benchmark the performance of their algorithms on ``in-the-wild" datasets. One of the sub-challenges of the EmotiW challenge is the audio-video emotion recognition sub-challenge, which is based on an augmented version of the AFEW dataset~\cite{Dhal12b} that contains short video clips extracted from movies that have been annotated for seven different emotions.  

Deep learning has played an important role in most of the sub-challenge winning submissions. In 2013, the winners presented a method that combines CNNs for static faces, an auto-encoder for human action recognition, a deep belief network for audio information, and a shallow network architecture for feature extraction of the mouth region~\cite{Kaho13}. The winners of the 2014 sub-challenge used CNNs for feature extraction of the aligned faces provided by the challenge organizers~\cite{Liu14}, while in 2016, the winners of the sub-challenge proposed a hybrid network architecture that combines 3D CNNs and a CNN-RNN in a late-fusion fashion~\cite{Fan16}.

While a variety of methods based on images have been proposed, the audio channel has been explored in a less extent. Existing approaches that exploit the audio channel for emotion recognition include the use of support vector machines (SVM)~\cite{Cortes:1995:SN:218919.218929,Fan16} , random forests~\cite{Breiman:2001:RF:570181.570182, Sun:2016:LDE:2993148.2997640} and CNNs trained on comprehensive acoustic vectors extracted by openSmile~\cite{Yan:2016:MFE:2993148.2997630}. We propose to fully exploit the  audio-channel information in this paper. Inspired by the recurrent support vector machines designed by Yun \textit{et al}.~\cite{Wang:2016:RSV:2911996.2912048} on event detection, we propose an LSTM~\cite{Hochreiter:1997:LSM:1246443.1246450} trained on short-term audio features extracted from segmented audio clips. Furthermore, a CNN-RNN network trained on image-like maps, formed by stacking short-term audio features, is also presented in this paper. The proposed hybrid network (Figure~\ref{fig1}) results in the combination of a CNN-RNN network trained on images and audio-based models, and achieves an overall accuracy of $55.61\%$ and $51.15\%$ on the validation and testing sets, respectively, surpassing the audio-video emotion recognition sub-challenge baseline of 38.81\% on the validation set with significant gains.

\section{The proposed method}
\label{sec:method}

\subsection{VGG-LSTM based on Faces}
A traditional CNN-LSTM neural network~\cite{Fan16,Sun:2016:LDE:2993148.2997640} is explored to learn emotion from faces. Video frames are extracted at a frequency of 60 fps. Faces and facial landmarks are first detected within each frame using the method described in~\cite{kazemi2014one}, then a 2D affine transformation where the left and right eye corners of all the images are aligned to the same positions is performed (the code of the face detection and alignment algorithms is developed based on~\cite{HHPE:CVPR15:frontalize}).

Aligned faces are used as input to the VGG-16 convolutional neural network~\cite{DBLP:journals/corr/SimonyanZ14a}. The VGG architecture is modified by changing the number of neurons in the last layer to 7, indicating 7 emotion classes. This modified VGG architecture is initialized with the parameters of the VGG-FACE model, except for the last fully-connected layer which is initialized with weights sampled from a Gaussian distribution of zero mean and variance $1\times 10^{-4}$, and trained from scratch with the learning rate for the weight and bias filters set to be 10 times larger than the overall learning rate. The VGG-FACE model was presented as the result of training the 16-layer VGG architecture on a large-scale dataset containing 2.6M images of 2.6K celebrities and public figures for face recognition in~\cite{Park15}

The training procedure is three-fold. First, the modified VGG network is trained on the facial expression recognition 2013 (FER-2013) database~\cite{DBLP:journals/fer2013}. The FER-2013 database contains $35887$ images corresponding to $7$ basic emotions. The idea of this step is to transfer the knowledge from face recognition to face emotion recognition. Second, the resulting model is fine-tuned on the detected faces of the AFEW dataset. Third, the layers of the fine-tuned model after the ``fc6" layer are replaced by a one-layer LSTM and a final fully connected layer with $7$ output units. The weights of the LSTM are initialized with values drawn from a uniform distribution over [-0.01, 0.01] and the bias terms are initialized to 0. The combined VGG-LSTM is trained end-to-end. The LSTM layer has $128$ nodes and the length of the input sequence is $16$. Face images extracted at every 8 frames of the original video sequence are selected as input to the VGG-LSTM network. Experimental results show that the frame gap helps improve the classification accuracy since facial change is more visible in this way. 


Unlike some existing works that first train the CNN and use their ``fc6" features as input vector sequences for the LSTM network, the proposed structure connects the VGG and LSTM networks end-to-end and learns all the parameters simultaneously. Experimental results show that our VGG-LSTM outperforms the results of the winner of the audio-video emotion recognition sub-challenge in 2016.   

\begin{table}
  \caption{Confusion matrix results of the VGG-LSTM network, trained on aligned faces, when tested on the validation set. The attained overall accuracy is $49.61\%$ and the unweighted average of the per-class accuracies is $46.66\%$.}
  \resizebox{\columnwidth}{!}{\begin{tabular}{|c|c|c|c|c|c|c|c|}
   \hline
     &AN&DI&FE&HA&NE&SA&SU\\
   \hline
    AN & 53.12 &6.25& 7.81 &0&17.19&3.12&12.50\\
       \hline
    DI & 17.50 &27.50& 7.50 &2.50&25.00&15.00&5.00\\
       \hline
    FE & 21.74 &4.35& 23.91 &13.04&6.52&17.39&13.04\\
       \hline
    HA & 7.94&1.59& 0 &84.13&0&4.76&1.59\\
       \hline
    NE & 11.11 &11.11& 7.94 &6.35&53.97&6.35&3.17\\
       \hline
    SA & 8.20 &4.92& 1.64 &6.56&22.95&55.74&0\\
       \hline
    SU & 21.74 &6.52& 17.39 &4.35&17.39&4.35&28.26\\
   \hline
\end{tabular}}
\label{tab1}
\end{table}

\begin{table}
  \caption{Confusion matrix results of the VGG-LSTM network, trained on the faces provided by the challenge organizers, tested on the validation set. The attained overall accuracy is $46.74\%$ and the unweighted average of the per-class accuracies is $43.17\%$.}
  \resizebox{\columnwidth}{!}{\begin{tabular}{|c|c|c|c|c|c|c|c|}
   \hline
     &AN&DI&FE&HA&NE&SA&SU\\
   \hline
    AN & 64.06 &1.56& 7.81 &1.56&12.50&6.25&6.25\\
       \hline
    DI & 22.50 &15.00& 5.00 &10.00&25.00&20.00&2.50\\
       \hline
    FE & 32.61 &8.70&26.09& 4.35 &13.04&8.70&6.52\\
       \hline
    HA & 9.52&3.17& 0 &73.02&6.35&6.35&1.59\\
       \hline
    NE & 14.29 &11.11& 1.59 &3.17&63.49&6.35&0\\
       \hline
    SA & 16.39 &11.48& 6.56 &8.20&13.11&40.98&3.28\\
       \hline
    SU & 32.61 &6.52& 17.39 &0&15.22&8.70&19.57\\
   \hline
\end{tabular}}
\label{tab2}
\end{table}

\begin{table}
  \caption{Confusion matrix of the audio SVM model on the validation set, with overall accuracy of $35.51\%$ and unweighted average of the per-class accuracies of $31.97\%$.}
  \resizebox{\columnwidth}{!}{\begin{tabular}{|c|c|c|c|c|c|c|c|}
   \hline
     &AN&DI&FE&HA&NE&SA&SU\\
   \hline
    AN & 76.56 &0& 3.12 &9.38&7.81&3.12&0\\
       \hline
    DI & 25.00 &0& 0 &42.50&20.00&12.50&0\\
       \hline
    FE & 23.91 &0&30.43& 23.91 &13.04&8.70&0\\
       \hline
    HA & 15.87&1.59& 9.52 &42.86&20.63&9.52&0\\
       \hline
    NE & 12.70 &1.59& 3.17 &34.92&46.03&1.59&0\\
       \hline
    SA & 11.48 &0& 11.48 &26.23&21.31&27.87&1.64\\
       \hline
    SU & 19.57 &0& 17.39 &36.96&13.04&13.04&0\\
   \hline
\end{tabular}}
\label{tab3}
\end{table}

\begin{table}
  \caption{Confusion matrix of the audio LSTM model on the validation set, with overall accuracy of $26.63\%$ and unweighted average of the per-class accuracies of $23.27\%$.}
  \resizebox{\columnwidth}{!}{\begin{tabular}{|c|c|c|c|c|c|c|c|}
   \hline
     &AN&DI&FE&HA&NE&SA&SU\\
   \hline
    AN & 48.44 &1.56& 0 &15.62&15.62&18.75&0\\
       \hline
    DI & 15.00 &2.50& 0 &35.00&32.50&15.00&0\\
       \hline
    FE & 30.43 &0&0& 19.57 &28.26&21.74&0\\
       \hline
    HA & 12.70&4.76& 0 &33.33&30.16&19.05&0\\
       \hline
    NE & 6.35 &3.17& 0 &17.46&52.38&20.63&0\\
       \hline
    SA & 11.48 &6.56& 0 &22.95&32.79&26.23&0\\
       \hline
    SU & 17.39 &0& 2.17 &30.43&36.96&13.04&0\\
   \hline
\end{tabular}}
\label{tab4}
\end{table}
\subsection{Acoustic SVM Classifier}

An SVM classifier, which is learned based on the 1582-dimensional acoustic features extracted using openSMILE, is incorporated into the hybrid network. Acoustic features include low level descriptors, such as energy, mel-frequency cepstral coefficients (MFCCs), linear predictive coding (LPC), zero-crossing rate (ZCR), spectral flux, spectral roll-off, chroma vector, and statistical features summarized by functions, such as mean and standard deviation. 

\subsection{LSTM based on Audio clips}
Instead of extracting holistic features, each audio is divided into segments of length 100 ms, using an overlapping factor of 50\%, and then segment-level features are extracted to form a sequence of vectors. Specifically, $34$ short-term features are extracted for each segment using pyAudioAnalysis~\cite{giannakopoulos2015pyaudioanalysis}. Features include ZCR, energy, entropy of energy, spectral centroid, spectral spread, spectral flux, spectral roll-off, MFCCs, chroma vector, and chroma deviation. Assume that the audio signal has length $m$, then the number of sequences $n$ is $(m-50)/50$. Therefore, this feature extraction process results in $n$ sequences of dimension $34\times1$. Since each audio is of different length, a sequence length converter is applied to make the number of sequences be at least $16$ by copying the last feature vector of the sequence $16-n$ times when the number of sequences is less than 16. A one-layer LSTM with $512$ neurons is trained on the sequence of feature vectors. Unlike the audio SVM model which focuses on the holistic properties of the signal, the audio LSTM model focuses on learning the dynamic temporal behavior of the audio signals.

\subsection{Inception(v2)-LSTM based on Audio Maps}

In this section, the sequence of feature vectors from Section 2.3 
is converted into sequential image-like maps. Specifically, the feature vectors are organized in matrix form to build an image-like map of dimension $34\times n$. The next step is to segment this image-like map into smaller maps of size $34\times34$ using an overlapping factor of $50\%$. For the architecture proposed in this section, the sequence length $n$ must be a multiple of $17$ and greater or equal than $34$. If this condition is not satisfied, then the last column of the $34\times n$ image-like map is replicated $n'-n$ times, where $n'$ is the closest multiple of 17 larger than $n$. This approach results in a sequence of $34\times34$ image-like maps, whose sequence length is $(n'-17)/17$. 


A network similar to the VGG-LSTM network described in section 2.1, Inception(v2)-LSTM, is developed to train on image-like maps. Instead of using the VGG architecture, we use Inception-v2~\cite{DBLP:journals/corr/SzegedyVISW15} to train the image-like maps first. The number of output units of the last layer is changed to $7$, and the training parameters, such as the learning rate and the weight decay are set the same as the ones used in Inception-v2 on ImageNet~\cite{imagenet_cvpr09}. 

After the training of the modified Inception-v2 on the individual image-like maps, the layers after the ``global\_pool" layer of the Inception-v2 architecture are replaced by a one-layer LSTM with $128$ neurons and a fully connected layer with $7$ outputs. The resulting network is referred to as Inception(v2)-LSTM. This network takes a sequence of 8 image-like maps at a time and learns the features end-to-end to model the dynamic temporal properties of the sequence. Since the sequence length of the image-like maps is $(n'-17)/17$ and needs to be greater than 8 to serve as input to the Inception(v2)-LSTM architecture, the last $34\times34$ image-like map of the sequence is copied $8-(n'-17)/17$ times to make the sequence satisfy the minimum length requirement. The initial learning rate is set to $0.001$ and decreases every $3000$ iterations by a factor of $10$. The batch size, the weight decay and the maximum number of iterations are set to $16$, $0.002$ and $10000$, respectively. 
\begin{table}
  \caption{Confusion matrix of the audio Incetion(v2)-LSTM model on the validation set, with overall accuracy of $26.63\%$ and unweighted average of the per-class accuracies of $23.14\%$.}
  \resizebox{\columnwidth}{!}{\begin{tabular}{|c|c|c|c|c|c|c|c|}
   \hline
     &AN&DI&FE&HA&NE&SA&SU\\
   \hline
    AN & 56.25 &0& 0 &29.69&10.94&3.12&0\\
       \hline
    DI & 12.50 &0& 0 &57.50&27.50&2.50&0\\
       \hline
    FE & 13.04 &0&2.17& 45.65 &36.96&2.17&0\\
       \hline
    HA & 11.11&0& 0 &52.38&26.98&9.52&0\\
       \hline
    NE & 6.35 &0& 0 &52.38&39.68&1.59&0\\
       \hline
    SA & 8.20 &0& 0 &63.93&16.39&11.48&0\\
       \hline
    SU & 6.52 &2.17& 2.17 &56.52&26.09&6.52&0\\
   \hline
\end{tabular}}
\label{tab5}
\end{table}

\begin{table}
  \caption{
  Confusion matrix results of submission 6 when evaluating the hybrid network on the testing dataset. The attained overall accuracy is $51.15\%$ and the un-weighted average of the per-class accuracies is $41.21\%$.}
  \resizebox{\columnwidth}{!}{\begin{tabular}{|c|c|c|c|c|c|c|c|}
   \hline
     &AN&DI&FE&HA&NE&SA&SU\\
   \hline
    AN & 77.55 &0& 4.08 &8.16&7.14&2.04&1.02\\
       \hline
    DI & 32.50 &10.00& 2.50 &12.50&20.00&20.00&2.50\\
       \hline
    FE & 31.43 &0&50.00& 1.43 &5.71&7.14&4.29\\
       \hline
    HA & 20.83&0& 1.39 &63.89&10.42&3.47&0\\
       \hline
    NE & 16.69 &1.04& 7.77 &6.22&50.78&11.92&2.59\\
       \hline
    SA & 22.50 &1.25& 11.25 &11.25&16.25&36.25&1.25\\
       \hline
    SU & 10.71 &3.57& 35.71 &10.71&14.29&25.00&0\\
   \hline
\end{tabular}}
\label{tab7}
\end{table}

\section{Experimental Results}
\label{sec:results}

\subsection{Database}
The AFEW database used in EmotiW 2017 contains 773, 383 and 653 audio-video movie clips in the training, validation and testing datasets, respectively. The task is to assign a single emotion label to a video clip from the $7$ basic emotions, namely, anger, disgust, fear, happiness, neutral, sad and surprise. Participants compete on the accuracy of the testing data\footnote{Note that since the class distribution is unbalanced, the accuracy participants compete on is the overall accuracy, which is computed on all the samples of the testing set. The unweighted average of the per-class accuracies is also provided in this paper.}.

\subsection{Results of the Proposed Models}
Confusion matrices for each model are shown in Tables~\ref{tab1} through ~\ref{tab5}. One VGG-LSTM model is trained on the aligned faces, which are obtained as described in Section 2.1, and another VGG-LSTM model is trained on the faces provided by the challenge organizers. Our best VGG-LSTM model achieves an overall classification accuracy of $49.61\%$, outperforming the $45.43\%$ accuracy obtained by the winner of the 2016 audio-video emotion recognition sub-challenge, which suggests that the frame gap introduced by the proposed VGG-LSTM model is a better way to represent the dynamics of face expression in video. The second VGG-LSTM model trained on the faces provided by the challenge organizers complements the proposed model, and the combination of the two achieves a classification accuracy of  $51.02\%$ on the validation set. 

Audio models, including audio SVM, audio LSTM and audio Inception(v2)-LSTM have lower accuracy than the VGG-LSTM models trained on faces. However, they perform well on the anger class, and therefore, it improves the overall accuracy of the hybrid network. 

The aforementioned deep models are combined using decision fusion. Grid search is employed to find the model weights that maximize the classification accuracy on the validation set. Fused hybrid network achieves a classification accuracy of $55.61\%$ on the validation set, while the challenge baseline is of accuracy $38.81\%$. When trained on a combination of the training and validation sets, the accuracy on the testing set of the proposed hybrid network is $51.15\%$. The corresponding confusion matrix is shown in Table~\ref{tab7}.

\section{Conclusions}
\label{sec:conclusions}
In this paper, we proposed an audiovisual-based hybrid network that combines the predictions of $5$ models for emotion recognition in the wild, with an emphasis on exploring audio channels. The overall accuracy of the proposed method achieves $55.61\%$ and $51.15\%$ classification accuracy on the validation and testing dataset, respectively, surpassing the audio-video emotion recognition sub-challenge baseline of 38.81\% on the validation set with significant gains.  

\bibliographystyle{IEEEbib}
\bibliography{main}

\end{document}